\documentclass{article}
\pdfoutput=1

\PassOptionsToPackage{numbers}{natbib}

     \usepackage[preprint]{neurips_2024}

\newcommand{\name}{{ReGS}\xspace}
\usepackage{xspace}

\makeatletter
\DeclareRobustCommand\onedot{\futurelet\@let@token\@onedot}
\def\@onedot{\ifx\@let@token.\else.\null\fi\xspace}
\def\eg{\emph{e.g}\onedot} 

\def\ie{\emph{i.e}\onedot}

\def\etal{\emph{et al}\onedot}
\newcommand{\printfnsymbol}[1]{%
	\textsuperscript{\@fnsymbol{#1}}%
}
\makeatother

\usepackage[utf8]{inputenc} 
\usepackage[T1]{fontenc}    
\usepackage{hyperref}       
\usepackage{url}            
\usepackage{booktabs}       
\usepackage{amsfonts}       
\usepackage{nicefrac}       
\usepackage{microtype}      
\usepackage{xcolor}         
\usepackage{graphicx}
\usepackage{tabularray}
\usepackage{amsmath}
\usepackage{comment}
\usepackage{enumerate}

\begin{document}
\title{\name: Reference-based Controllable Scene Stylization with Gaussian Splatting}

%

\author{%
  Yiqun~Mei\thanks{Equal contribution} \quad
  Jiacong~Xu\printfnsymbol{1} \quad Vishal M.~Patel \\
  Johns Hopkins University \\\texttt{\{ymei7,jxu155, vpatel36\}@jhu.edu}}

\maketitle

\begin{abstract}
  Referenced-based scene stylization that edits the appearance based on a content-aligned reference image is an emerging research area. Starting with a pretrained neural radiance field (NeRF), existing methods typically learn a novel appearance that matches the given style. Despite their effectiveness, they inherently suffer from time-consuming volume rendering, and thus are impractical for many real-time applications. In this work, we propose \name, which adapts 3D Gaussian Splatting (3DGS) for reference-based stylization to enable real-time stylized view synthesis. Editing the appearance of a pretrained 3DGS is challenging as it uses \textit{discrete} Gaussians as 3D representation, which tightly bind appearance with geometry. Simply optimizing the appearance as prior methods do is often insufficient for modeling continuous textures in the given reference image. To address this challenge, we propose a novel texture-guided control mechanism that adaptively adjusts local responsible Gaussians to a new geometric arrangement, serving for desired texture details. The proposed process is guided by texture clues for effective appearance editing, and regularized by scene depth for preserving original geometric structure. 
With these novel designs, we show \name can produce state-of-the-art stylization results that respect the reference texture while embracing real-time rendering speed for free-view navigation. 
\end{abstract}

\section{Introduction}\label{intro}
Stylizing a 3D scene based on a 2D artwork is an active research area in both computer vision and graphics \cite{huang2021learning, mu20223d, stylizednerf, snerf, nerf-art, hypernetwork, ins}.
%
\begin{figure}
	\centering
	\includegraphics[width=1\textwidth]{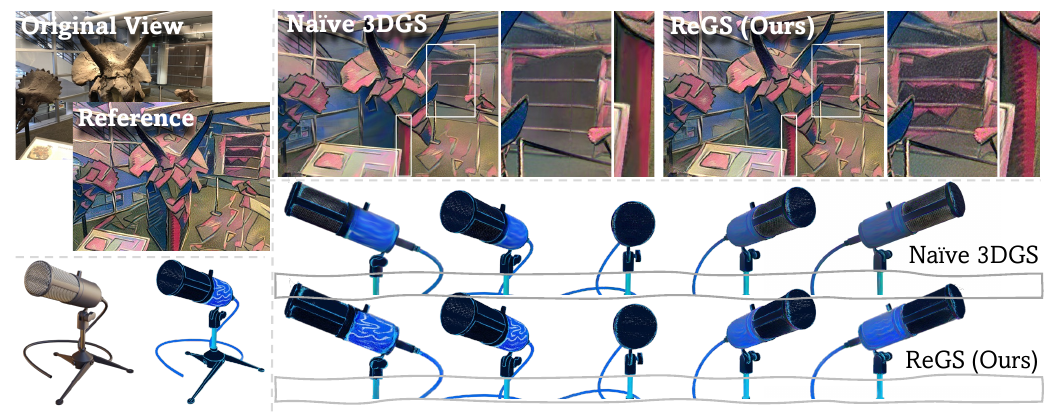}
\vspace{-6mm}	
\caption{Given a pretrained 3DGS model of the target scene and its paired style reference, \name enables real-time stylized view synthesis (at 134 FPS) with high-fidelity texture well-aligned with the reference. In contrast, only optimizing the appearance of 3DGS (denoted as Naive 3DGS), as previous methods~\cite{arf,stylerf,stylizednerf,ref-npr,hypernetwork} do, fails to capture many texture details in the reference. We tackle the challenges in high-fidelity appearance editing with a texture-guided control mechanism that is significantly more effective than the default density control~\cite{3dgaussiansplatting} in addressing texture underfitting. Side-by-side comparisons with default density control can be found in Figure~\ref{fig:control}.}  
 \vspace{-6mm}\label{workflow}
\end{figure}
One important direction of stylization aims to precisely stylize the scene appearance based on a 2D content-aligned reference image drawn by users~\cite{ref-npr}. Such problem has numerous applications in digital art, film production and virtual reality.
In the classical graphics pipeline, completing this task requires experienced 3D artists to manually create a UV texture map as input to the shader, a tedious process requiring professional knowledge, significant time, and effort.

Over the past decades, tremendous progress has been made in automatic scene stylization by leveraging view synthesis methods.
While early attempts \cite{huang2021learning, mu20223d,hollein2022stylemesh,fivser2016stylit} suffer from geometry errors of point clouds or meshes, more recent methods \cite{stylerf, arf, stylizednerf, snerf, nerf-art, hypernetwork, ins} rely on radiance field (NeRF) \cite{nerf}, a powerful implicit 3D representation, to deliver high-quality renditions that are perceptually similar to the reference image. 
A typical stylization workflow starts from a pretrained NeRF model of the target scene, followed by an appearance optimization phase to match the given style. The density function is always fixed to maintain the scene geometry~\cite{arf,stylerf,stylizednerf,ref-npr,hypernetwork} . 
Despite their promising results, NeRF-based approaches consume high training and rendering costs in order to obtain satisfactory results. 
Although some recent efforts make fast training possible \cite{plenoxels, dvgo, tensorf, nsvf, autoint, instantngp}, the improvement in efficiency often comes at the price of degraded visual quality. Meanwhile, real-time rendering at inference time still remains challenging. 

Recently, 3D Gaussian Splatting (3DGS) \cite{3dgaussiansplatting} has become an emerging choice for representing 3D scenes. 
3DGS creates millions of colored Gaussians with learnable attributes to jointly represent the target scene geometry and appearance.
Importantly, it adopts splatting-based rasterization \cite{zwicker2001ewa} to replace the time-consuming volume rendering of NeRF models, providing remarkably faster rendering speed while maintaining comparable visual quality. However, as it uses \textit{discrete} 3D Gaussians to represent the reconstructed scene, optimizing their appearance with a fixed geometry layout (as NeRF-based methods do) is often inadequate to capture the \textit{continuous} texture variance in the reference image.
This ``appearance-geometry entanglement" makes applying 3DGS to applications that require novel appearance, \ie stylization, challenging. For 3DGS, how to properly control and edit the appearance without distorting the original geometry remains under-explored.   

In this paper, we present a novel reference-based scene stylization method using 3DGS, dubbed \name, to enable real-time stylized view synthesis with high-fidelity textures well-aligned with the given reference. Similar to previous methods, our approach starts with a pretrained 3D Gaussian model of the target scene. The core enabler of \name is a novel texture-guided control procedure that makes high-fidelity appearance editing with ease. In particular, we adaptively adjust the local arrangement of responsible Gaussians in the appearance underfitting regions to a state that the desired textures specified in the reference image can be faithfully expressed. The control process is designed to \textbf{(1)} automatically identify target local Gaussians using texture clues, and \textbf{(2)} structurally distribute tiny Gaussians for fast detail infilling while \textbf{(3)} sticking to the original scene structure via a depth-based regularization. With these novel designs, \name is able to learn consistent 3D appearance that accurately follows the given reference image.

Following~\cite{ref-npr}, we train \name on a set of pseudo-stylized images for view consistency, which are synthetic multi-view data created using extracted scene depth, alongside with a template-based matching loss to ensure style spread to the occluded regions. By combining these techniques with the proposed texture-guided control, \name is capable of producing visually appealing stylization results that attain both geometric and perceptual consistency. Through extensive experiments, we demonstrate that \name achieves state-of-the-art visual quality compared to existing stylization methods while enabling real-time view synthesis by embracing the fast rendering speed of Gaussian Splatting.

\section{Related Work}
\label{sec:related_work}
\subsection{3D Scene Representation}
\noindent\textbf{Neural Radiance Field.} Reconstructing 3D scene from multi-view collections is a long-standing problem in computer vision. Early approaches adopting explicit mesh~\cite{waechter2014let,debec1996modeling,liu2019soft,wang2023creative} or voxel~\cite{kutulakos2000theory,seitz1999photorealistic,szeliski1998stereo} based representations often suffer from geometry error and lack of appearance details \cite{nerfreview}. Recent methods \cite{mipnerf, mipnerf360, nerfw, pixelnerf, refnerf, hdrnerf} adopt learnable radiance fields \cite{nerf} to capture 3D scene implicitly and outperform previous techniques by a large margin. However, NeRF models require millions of network queries for a single rendition that can be extremely time and resource-consuming. To reduce the training time, advanced methods adopt explicit/hybrid representations including voxel grid~\cite{nsvf, dvgo, plenoxels,sun2022direct, baking}, octree~\cite{plenoctrees, wang2022fourier, bai2023dynamic}, planes~\cite{chan2022efficient,cao2023hexplane,tensorf,fridovich2023k} and hash grid~\cite{instantngp}, and successfully reduce the training time from days to minutes. Nevertheless, the rendering speed at inference time is still limited by their volumetric nature, which requires dense sampling along a ray to generate a single pixel. 

\medskip
\noindent\textbf{3D Gaussian Splatting}. Recently, 3D Gaussian Splatting (3DGS)~\cite{3dgaussiansplatting} achieves real-time novel view synthesis based on a differentiable rasterizer~\cite{zwicker2001ewa} that efficiently projects millions of 3D Gaussians to a 2D canvas. Given its high efficiency, 3DGS becomes a promising solution to enable real-time vision applications, such as human avatar \cite{hu2023gaussianavatar, li2023animatable, li2024gaussianbody, kocabas2023hugs}, 3D object and immersive scene creation \cite{tang2024dreamgaussian, yi2023gaussiandreamer, chung2023luciddreamer, liu2023humangaussian}, relighting \cite{gao2023relightable, saito2023relightable, liang2023gs, shi2023gir}, surface or mesh reconstruction \cite{guedon2023sugar, chen2023neusg}, 3D segmentation \cite{zhou2023feature, cen2023segment, dou2024cosseggaussians}, and SLAM \cite{yugay2023gaussian, matsuki2023gaussian, li2024sgs}. Motivated by its high efficiency, our work explores 3DGS to enable real-time stylized view navigation.

\subsection{2D Stylization} 
\noindent\textbf{Arbitrary Style Transfer.} Our method is related to the general 2D stylization~\cite{jing2019neural}, which transfers the style from an artwork to a target image while maintaining the original content structure. In the pioneering work, Gatys \etal \cite{style_cnn} introduce an iterative scheme that progressively reduces the difference between the Gram statistics of generated image and style image features, yet lengthy optimization is required per style. To improve efficiency, later methods \cite{linear_style_transfer, adain, adaattn, sanet, din} focus on arbitrary image/video stylization by transferring the content image to target style spaces in a zero-shot manner. For example, Huang \etal \cite{adain} introduce AdaIN, which achieves real-time stylization by matching content features with the mean and standard deviation of style features. Linear style transfer~\cite{linear_style_transfer} instead predicts a linear transformation matrix based on both content and style pairs. For video stylization, it is crucial to maintain temporal coherence of the stylized frames. Techniques \cite{video0, video1, video2, video3, wu2020preserving, deng2021arbitrary}, such as flow-based wrapping \cite{video1}, global SSIM constraint \cite{wu2020preserving}, and inter-frame feature similarity \cite{deng2021arbitrary}, are proposed to ensure the consistency. 

\medskip
\noindent\textbf{Optimization-based Style Transfer.} While arbitrary style transfer is desirable in terms of flexibility, they often fall short of reproducing small stylistic patterns and lack high-frequency details \cite{arf, nnst}. Optimization-based Stylization \cite{li2016combining, risser2017stable, gu2018arbitrary,kolkin2019style,liao2017visual, nnst} is still the primary choice to ensure visual quality. For instance, a coarse-to-fine strategy is proposed by Liao \etal \cite{liao2017visual} to compute the nearest-neighbor field and build a semantically meaningful mapping between input and style images for visual attribute transfer. Kolkin \etal \cite{nnst} reach state-of-the-art stylization quality by replacing the content features with the nearest style feature.  To enable better controllability, example-based methods \cite{jamrivska2019stylizing, selim2016painting, shih2014style, texler2020interactive} perform wrapping or stylizing based on the aligned correspondences between the style reference and content images. However, their 2D alignment is generally unsuitable for 3D scenes due to occlusions, leading to flickering effects \cite{ref-npr}.

\subsection{3D Stylization} 
3D scene stylization extends artistic works beyond the 2D canvas~\cite{chen2023advances}. Early works \cite{huang2021learning, mu20223d} typically back-project image colors as 3D point cloud for processing, and project stylized point features back to 2D for view synthesis. Yet, using point cloud often fails to represent complicated geometry and produces artifacts for complex scenes~\cite{arf}. 

Benefiting from NeRF, methods stylizing radiance fields \cite{stylerf, arf, stylizednerf, snerf, nerf-art, hypernetwork, ins} have shown visually compelling and geometry-consistent results than previously possible. Similar to image stylization, several works~\cite{stylerf,stylizednerf, hypernetwork,ins} deal with arbitrary or multiple style transfer using various techniques such as 2D-3D mutual learning~\cite{stylizednerf}, deferred style transformation~\cite{stylerf}, and hypernetwork \cite{hypernetwork}. 
While a universal stylizer might be desirable, these methods can only transfer the overall color tone and lack detailed style patterns, \ie brushstrokes. Per-style optimization is still required for better visual quality. Among these methods \cite{snerf, arf, nerf-art,zhang2023transforming}, ARF \cite{arf} shows state-of-the-art stylization capability by progressively matching the generated features with the closest style feature via nearest neighbor search. However, these methods are designed for transferring styles from an arbitrary reference and lack controllability over generated results.  To this end, Ref-NPR \cite{ref-npr} introduces a reference-based scheme that controls stylized appearance based on a content-aligned reference image. Our work also focuses on this setting.

\section{Method}
\begin{figure}[t]
	\centering
	\includegraphics[width=1\textwidth]{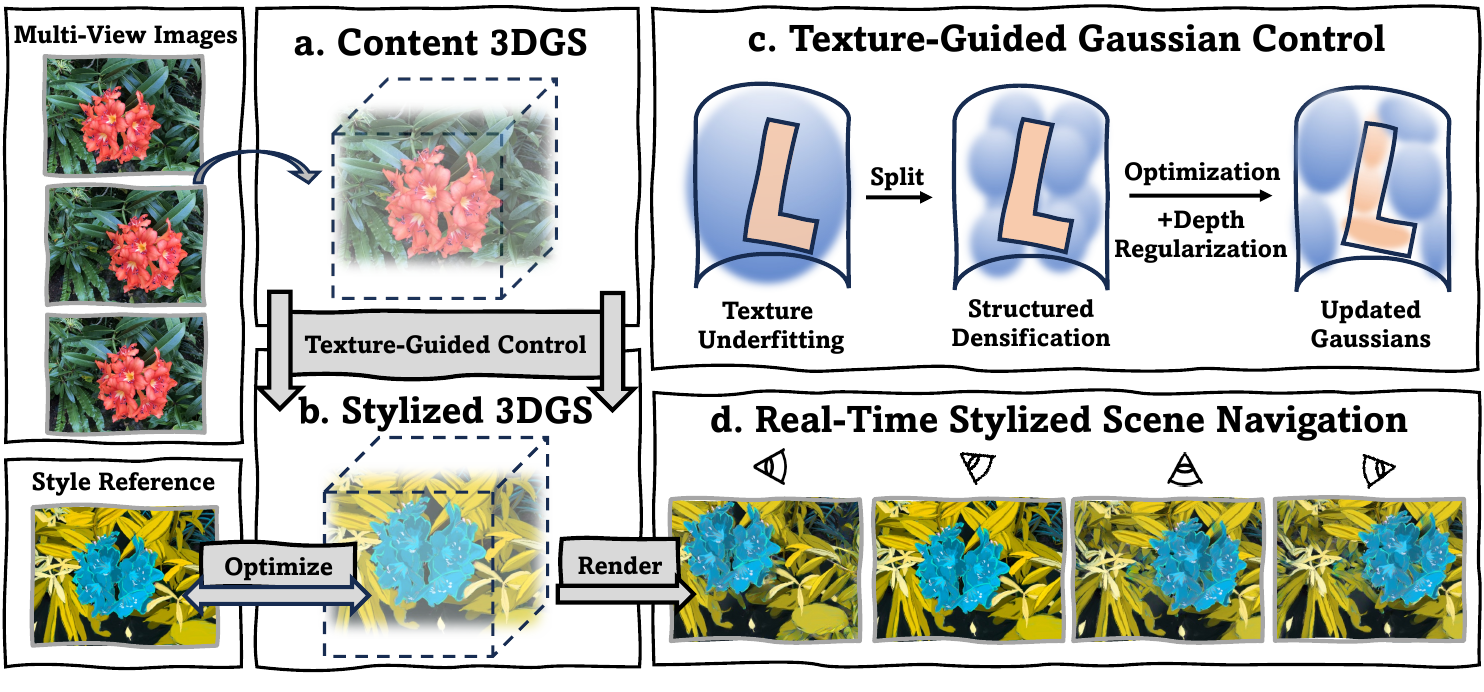}
\vspace{-6mm}	
 \caption{\textbf{An overview of \name}. (a) The proposed method starts with a pretrained content 3DGS of the target scene, and (b) outputs a stylized 3DGS that follows the reference. (c) We propose Texture-Guided Gaussian Control that can progressively resolve texture underfitting by automatically locating responsible Gaussians and adjusting local geometry layout for fitting high-frequency textures. (d) Once training is done, our method enables real-time stylized scene navigation.}
 \vspace{-6mm}\label{workflow}
\end{figure}

An overview of \name is shown in Figure~\ref{workflow}. \name takes a pretrained 3DGS model (Figure~\ref{workflow} (a)) of the target scene as well as a content-aligned reference image as inputs. It outputs a stylized 3DGS model (Figure~\ref{workflow} (b)) that bakes the texture of the reference image into the scene and enables real-time stylized views synthesis (Figure~\ref{workflow} (d)). 

As 3DGS represents a scene as discrete Gaussians, simply optimizing its appearance often cannot capture the continuous texture details in the reference image. We introduce a texture-guided control mechanism to progressively address this challenge (Sec.~\ref{densify}). To ensure no geometry distortion happens during optimization, we propose a geometry regularization using scene depth (Sec.~\ref{depth}). We then introduce two techniques to encourage perceptual-consistent renditions (Sec.~\ref{pesudo_view}). Finally, we describe our training objectives in Sec.~\ref{loss}.

\subsection{Preliminary: 3D Gaussian Splatting} \label{prel}
Before introducing our method, we first provide a brief review of 3D Gaussian Splatting~\cite{3dgaussiansplatting}. 3DGS represents the scene explicitly by a collection of learnable Gaussians. Each 3D Gaussian is attributed by a positional vector $\mathbf{\mu} \in \mathbb{R}^{3}$ and a 3D covariance matrix $ \mathbf{\Sigma} \in \mathbb{R}^{3\times 3}$. Its influence on a space point $\mathbf{x}$ is proportional to a Gaussian distribution:
 \setlength{\belowdisplayskip}{2pt} \setlength{\belowdisplayshortskip}{2pt}
\setlength{\abovedisplayskip}{2pt} \setlength{\abovedisplayshortskip}{2pt}
\begin{equation}
    G(\mathbf{x}) = e^{-\frac{1}{2}(\mathbf{x}-\mathbf{\mu})^{\top}\mathbf{\Sigma}^{-1}(\mathbf{x}-\mathbf{\mu})}.
\end{equation}
 By definition, the covariance matrix should be positive semi-definite. This is achieved by decomposing $\mathbf{\Sigma}$ into a scaling matrix $\mathbf{S}$ and a quaternion $\mathbf{R}$ \ie $
    \mathbf{\Sigma} = \mathbf{R}\mathbf{S}\mathbf{S}^{\top}\mathbf{R}^{\top}$. 
 Each Gaussian also stores an opacity value $\alpha_{i}$ and a view-dependent color represented by Spherical Harmonic (SH) coefficients. 
    
The rendering procedure is implemented as splatting-based rasterization~\cite{zwicker2001ewa} which projects Gaussians to a 2D canvas. The projected 2D splats are then sorted based on the depth to the camera. After sorting, the final color for each pixel is computed through $\alpha$-blending:
\begin{equation}
C=\sum_{i=1}^{n}c_{i}\alpha_{i}^{\prime}\prod_{j=1}^{i-1}(1-\alpha_{j}^{\prime})\label{eq: blend},
\end{equation}
where $c_{i}$ is a view-dependent color of the $i$-th Gaussian computed from SH. $\alpha_{i}^{\prime}$ is the multiplication result of the learned opacity $\alpha_{i}$ and evaluated value of the projected 2D Gaussian. 

During optimization, heuristic controls are employed to adaptively manage the density of Gaussians to better represent the scene. Specifically, it densifies Gaussians with large positional gradients to capture missing geometry and prunes Gaussians with small opacity to improve compactness.

\subsection{Texture-Guided Gaussian Control}~\label{densify}
As a discrete scene representation, the geometry layout and arrangement of Gaussians essentially limit the range of appearance it can express. For example, as shown in Figure~\ref{workflow} (c), appearance underfitting happens frequently at the area where local granularity of Gaussians is greater than the variance of the texture, \eg a smooth colored surface in the original scene is painted with rich details in the reference view. \name addresses such challenges via a novel texture-guided control that splits these responsible local Gaussians into a denser set suitable for high-frequency texture. Specifically, the proposed mechanism automatically identifies responsible Gaussians using texture clues and structurally replaces them with a denser set of tiny Gaussians to compensate for the missing details. We describe important designs of the proposed algorithm below.

\medskip
\noindent{\textbf{Texture Guidance.}} The control algorithm is directly guided by texture clues. Specifically, we
accumulate color gradients of all Gaussians over iterations and select Gaussians with larger gradient magnitude than a threshold for densification. We found that a larger color gradient corresponds to Gaussians that have large texture errors while the optimization struggles to find the correct colors to reduce the loss. This control scheme shares a similar spirit with the original control scheme in 3DGS, where they leverage positional gradients to locate Gaussians responsible for missing geometric features. But in stylization, scene geometry is already well-reconstructed through pretraining, and therefore, the positional gradient is no longer informative. As demonstrated in Figure~\ref{fig:control}, our color-based control scheme is more sensitive for pinpointing Gaussians with missing textures than the positional-based solution. In practical implementation, we increase density based on the gradient statistics of every 100 iterations.

\smallskip
\noindent{\textbf{Structured Densification.}} Traditional mesh subdivision~\cite{zorin1996interpolating} cuts large faces into more sub-faces to express surface details. Sharing a similar spirit, we structurally split each responsible Gaussians into a structured denser set to better represent texture details. Intuitively, after densification, newly added Gaussians need to approximate the original space coverage to avoid inducing large geometry errors, and they should be sufficiently small and form a dense set to capture appearance variance. Based on these considerations, we propose a structured densification scheme that adds tiny Gaussians into the most representative locations surrounding their parent Gaussian. Specifically, we use nine tiny Gaussians to replace a parent Gaussian. Eight of them correspond to eight separate octants divided by the equatorial plane and perpendicular meridian planes of the original ellipsoid. And the rest is placed at the original center. We reduce their size by shrinking the scales with a factor of 8 and copy remaining parameters from their parent Gaussian. We empirically found this setup can roughly maintain a space coverage that approximates the original geometry. As optimization continues, the densified Gaussians are progressively updated to infill missing textures.
\begin{figure}[t]
	\centering
	\includegraphics[width=1\textwidth]{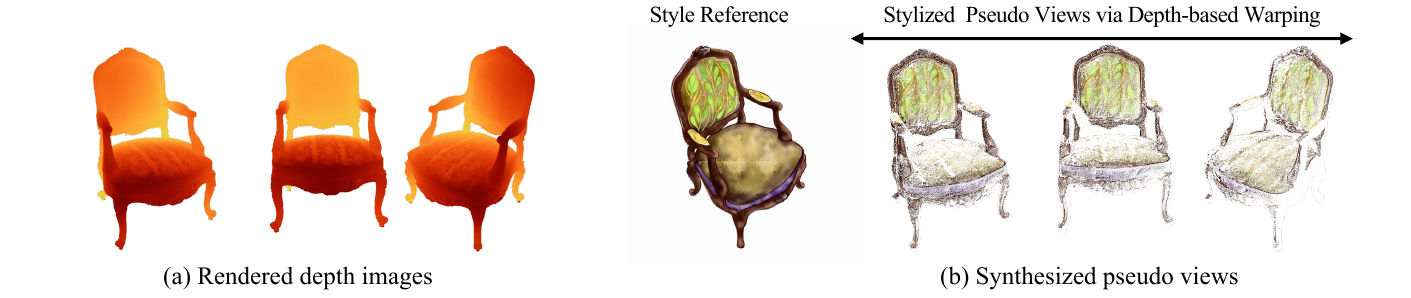}
\vspace{-7mm}	
 \caption{Examples of (a) rendered depth maps using Eq.\ref{depth_eq} and (b) synthesized stylized pseudo views.}
 \vspace{-7
 mm}\label{fig:depth}
\end{figure}
\subsection{Depth-based Geometry Regularization}~\label{depth}
While our control mechanism progressively improves texture details, it is essential to ensure the original scene geometry is preserved after optimization. We resort to the scene depth as an additional regularization to penalize geometry changes. Examples of rendered depth are shown in Figure~\ref{fig:depth} (a). Formally, we derive the scene depth via a $\alpha$-blending-based equation:
\begin{equation} \label{depth_eq}
    d = \sum_{i=1}^{n} d_{i} \alpha_{i}' \prod_{j=1}^{i-1}(1-\alpha_{j}'),
\end{equation} 
where the $d_{i}$ is the z-buffer associated with the $i$th Gaussian and $\alpha_{i}{'}$ is the same evaluated opacity in Eq.~\ref{eq: blend}. $d_{i}$ is computed by projecting the 3D location $\mu$ to the camera space.

The depth regularization is defined as the $L_{1}$ distance between a depth image $D_{i}$ rendered from original scene model $\mathbf{m}$ and a depth image $\widehat{D}_{i}$ rendered from the stylized model $\widehat{\mathbf{m}}$ using the same camera pose $\phi_{i}$ \ie $\mathcal{L}_{depth} =\| \widehat{D}_{i}-D_{i}\|_{1}$. 

\subsection{View-Consistent Stylization}~\label{pesudo_view}
For stylization, it is necessary to ensure the stylized appearance is consistent across different viewpoints and inpaints the occluded areas. Following~\cite{ref-npr}, we adopt two strategies to address them. 

\smallskip
\noindent{\textbf{Stylized Pseudo View Supervision.}} An image with paired depth contains sufficient information to re-render from nearby viewpoints~\cite{mark1997post}. This allows us to create a set of stylized pseudo views for obtaining additional supervision from the reference image. Our synthesis approach is very similar to classic depth-based 3D warping~\cite{mark1997post,mcmillan2023plenoptic}. Specifically, we back-project the reference image $S_{R}$ to the world space using the depth image $D_{R}$ and its camera pose ${\phi_{R}}$. Then, we re-project these 3D points back to a new viewpoint $\phi_{i}$. The resulting 2D image ${S}_{i}$ is used as an additional style supervision. Examples of the created pseudo views are shown in Figure~\ref{fig:depth} (b). 
It is important to make sure supervision only happens on meaningful pixels, \ie they are projections of 3D points that are visible from the current viewpoint $\phi_{i}$. Therefore, we conduct a visibility check by comparing the depth between the 2D projections of the 3D points and the depth image $D_{i}$ from the current viewpoint $\phi_{i}$. This results in a visibility mask $M_{i}$. Given the pseudo views and visibility masks, one can define a pseudo view supervision loss as 
\begin{equation}
\mathcal{L}_{view} = \frac{1}{\|M_{i}\|_{0}}\|M_{i}\widehat{S}_{i} - M_{i}S_{i}\|_{1},
\end{equation}
where ${\|.\|_{0}}$ is the $\ell_{0}$-norm that counts the number of valid pixels and $\widehat{S}_{i}$ is renderings of the stylized model $\widehat{\mathbf{m}}$.

\smallskip
\noindent{\textbf{Template Correspondence Matching (TCM) Loss.}} To ensure stylized appearance spreads to the occluded areas, we adopt the same TCM loss proposed in~\cite{ref-npr}. We briefly describe it here and refer readers to~\cite{ref-npr} for more details. TCM regularizes the difference of semantic correspondences before and after stylization. Given the style reference $S_{R}$, its corresponding view $I_{R}$, and a scene image $I_{i}$ rendered from a camera pose $\phi_{i}$, it constructs a guidance feature $F_{G}$ by
    $F_{G_{i}}^{(x,y)} = F_{S_{R}}^{(x^{*}, y^{*})} $
where 
\begin{equation}
    {(x^{*}, y^{*})}=\operatorname*{argmin}_{x
    ',y'} \textbf{dist}(F_{I_{i}}^{(x,y)}, F_{I_{R}}^{(x', y')})\label{eq: feature_idx}.
\end{equation}
  Here, $F_{S_{R}}$, $F_{I_{R}}$, $F_{I_{i}}$ denote deep semantic features of image $S_{R}$ $I_{R}$, and $I_{i}$ extracted by an ImageNet pretrained VGG~\cite{simonyan2014very}. \textbf{dist} denotes the cosine distance. After obtaining the guidance feature, TCM loss is defined as a cosine distance loss:
\begin{equation}
    \mathcal{L}_{TCM} = \textbf{dist}(F_{\widehat{s}_{i}}, F_{G_{i}}),\label{eq:dist}
\end{equation}
where $F_{\widehat{s}_{i}}$ is the extracted VGG features of the generated stylized view $\widehat{S}_{i}$.

\subsection{Training Objectives}~\label{loss}
Besides aforementioned depth loss $\mathcal{L}_{depth}$, pseudo view supervision loss 
$\mathcal{L}_{view}$ and TCM loss $\mathcal{L}_{TCM}$, \name further optimizes a reconstruction loss $\mathcal{L}_{rec}$ and a coarse color-matching loss $\mathcal{L}_{color}$~\cite{ref-npr}. The reconstruction loss is defined as the $L_{1}$ distance between the reference $S_{R}$ and the corresponding stylized output $\hat{S}_{R}$ to enforce appearance baking. The color-matching loss is defined as
\begin{equation}
\mathcal{L}_{color} = \|\widehat{S}^{(x,y)}_{i} - S_{R}^{(x^{*},y^{*})}\|_{2}^{2}, 
\end{equation}
where $S^{(x,y)}$ denotes the average color of a patch associated with feature-level index ${(x, y)}$. Feature-level index is computed using Eq.~\ref{eq: 
feature_idx}. This loss is directly adapted from~\cite{ref-npr} to encourage overall color matching in the occluded area. The overall loss for \name can be expressed as 
\begin{equation}
    \mathcal{L} = \lambda_{rec} \mathcal{L}_{rec} + \lambda_{depth}\mathcal{L}_{depth} + \lambda_{view}\mathcal{L}_{view} + \lambda_{tcm} \mathcal{L}_{TCM} + \lambda_{color}\mathcal{L}_{color}
\end{equation}
where $\lambda_{(.)}$ denotes the balancing parameter.

\subsection{Implementation and Training Details} \label{impl_detail}
\name uses 3D Gaussians~\cite{3dgaussiansplatting} as the scene representation and is built upon their official codebase. We follow the default parameter settings to obtain the pretrained 3D Gaussian model of the photo-realistic scene. For stylization, as we do not expect view-dependent effects, we discard the higher order SH and only render diffuse color in the stylization phase. Therefore, content images used in $\mathcal{L}_{TCM}$ and $\mathcal{L}_{color}$ are the results of this diffuse model.

For texture-guided control, we start accumulating gradients after a warm-up of 100 iterations and then perform the densification operation based on the color gradient statistics of every 100 iterations. The control process stops when it reaches half of the total iterations. The gradient threshold is empirically set to $1e-5$ at the beginning, and we linearly reduce it to $5e-6$ to allow for refining tiny details in the later training stage. Following~\cite{ref-npr, arf}, we use  the ImageNet pretrained VGG16~\cite{simonyan2014very} as the feature extractor and use the features produced by \textit{relu\_}3 and \textit{relu\_}4 in $\mathcal{L}_{TCM}$. For balancing parameters we set $\lambda_{rec}=\lambda_{tcm}=1$, $\lambda_{depth}=10$, $\lambda_{view}=2$, and $\lambda_{color}=15$. At each iteration, we always sample two views: the reference view and a random view. We train our model for 3000 iterations. The proposed method is implemented using PyTorch and trained on one A5000 GPU.

\section{Experiments}
In this section, we demonstrate the stylization quality and our designs through extensive experiments. \textbf{More experiment results and ablation} can be found in the supplemental file and accompanied video.

\subsection{Datasets} The only available reference-based stylization dataset is provided by ~\cite{ref-npr}. The dataset contains 12 selected scenes from  {Blender~\cite{nerf}}, {LLFF~\cite{locallightfieldfusion}}, and {Tanks and Temples~\cite{Knapitsch2017}}. Each scene is paired with a content-aligned reference image.

\begin{figure}[t!]
	\centering
	\includegraphics[width=1\textwidth]{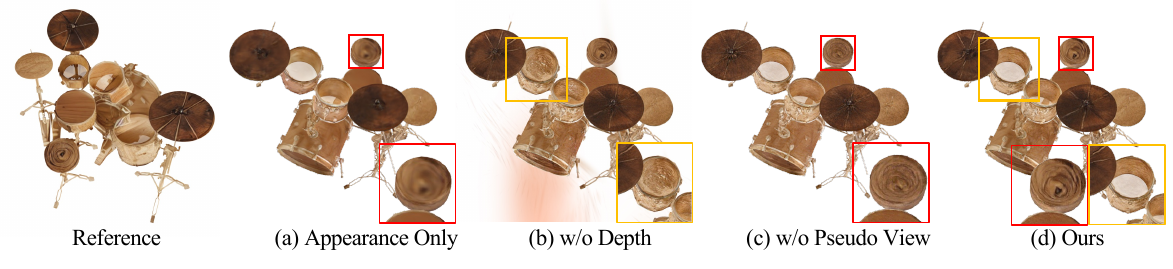}
\vspace{-7mm}	
\caption{\textbf{Ablation study on different components of \name}. (a) Optimizing only the appearance of a 3DGS model cannot reproduce texture details. (b) Removing depth regularization causes Gaussians to float out from the surface and distort the origin geometry. (c) Without pseudo-view supervision, results lack view consistency. (d) Our full model produces the best results that faithfully respect the texture in the reference.}
 \vspace{-5mm}\label{fig:ab}
\end{figure}

\begin{figure}[t]
	\centering
	\includegraphics[width=1\textwidth]{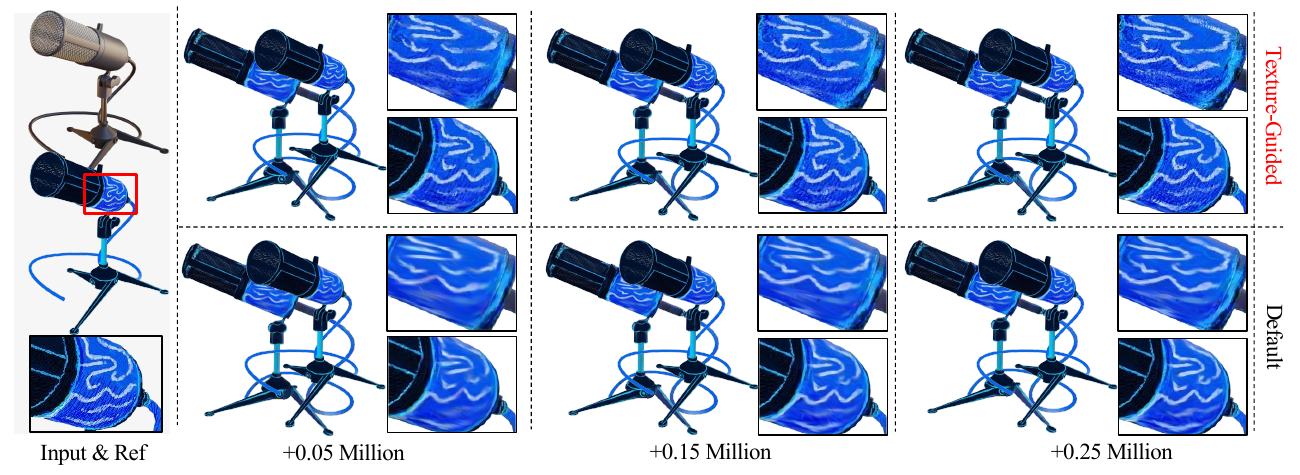}
\vspace{-7mm}	
 \caption{\textbf{Effectiveness of Texture-Guided Control.} We conduct controlled experiments by limiting the number of newly densified Gaussians throughout optimization. The pretrained model contains 0.3M Gaussians. The proposed texture-guided control can more faithfully reproduce the target texture details with a small number of Gaussians added (0.05M). The default strategy struggles to capture high-frequency details, even with a large number of Gaussians added (0.25M).}
 \vspace{-4mm}\label{fig:control}
\end{figure}

\vspace{-1mm}
\subsection{Ablation Study}
We conduct controlled experiments to analyze the effectiveness of each design choice in \name. Results are illustrated in Figures~\ref{fig:ab} \&~\ref{fig:control}.
As illustrated in Figure~\ref{fig:ab}, replacing any components of \name will harm the stylization quality. For example, Figure~\ref{fig:ab} (a) shows that optimizing only the appearance with fixed geometry arrangement like previous methods~\cite{arf,hypernetwork,stylizednerf,stylerf,ref-npr} do fails to recover the texture details. As shown in Figure~\ref{fig:ab} (b), after removing depth regularization, Gaussians float out from the surface and distort the original scene geometry. Similarly, discarding the pseudo view supervision will induce view-inconsistency as highlighted in the inset (Figure~\ref{fig:ab} (c)). The full model overcomes these issues and produces more visually appealing results that follow the given reference.

\medskip
\noindent\textbf{Effectiveness of Texture-Guided Control.} The core enabler of \name is the proposed texture-guided control mechanism that makes high-fidelity appearance editing with ease. Here, we demonstrate its effectiveness by comparing it with the default positional-gradient-guided density control~\cite{3dgaussiansplatting} in addressing {texture underfitting}. Specifically, we conduct controlled experiments by setting a series of limits on the total number of Gaussians that can grow throughout optimization. Results are reported in Figure~\ref{fig:control}.
One can see that by growing a very small amount of Gaussians (0.05M), the proposed texture-guided method can quickly infill most of the missing details. With more Gaussians added, it can further faithfully reproduce the given texture. In contrast, even with a large amount of new Gaussians (0.25M) created, the default method can barely capture high-frequency texture details. This is mainly because positional gradient is not sensitive to texture errors. As such, it fails to grow Gaussians in the regions with texture underfitting. And further moving these incorrectly placed Gaussians to the correct place for texture infilling is challenging. These results demonstrate our method is indeed more favorable for addressing appearance underfitting. \textbf{Study on structured densification can be found in the supplement.}

\begin{figure*}[t]
	\centering
	\includegraphics[width=1\textwidth]{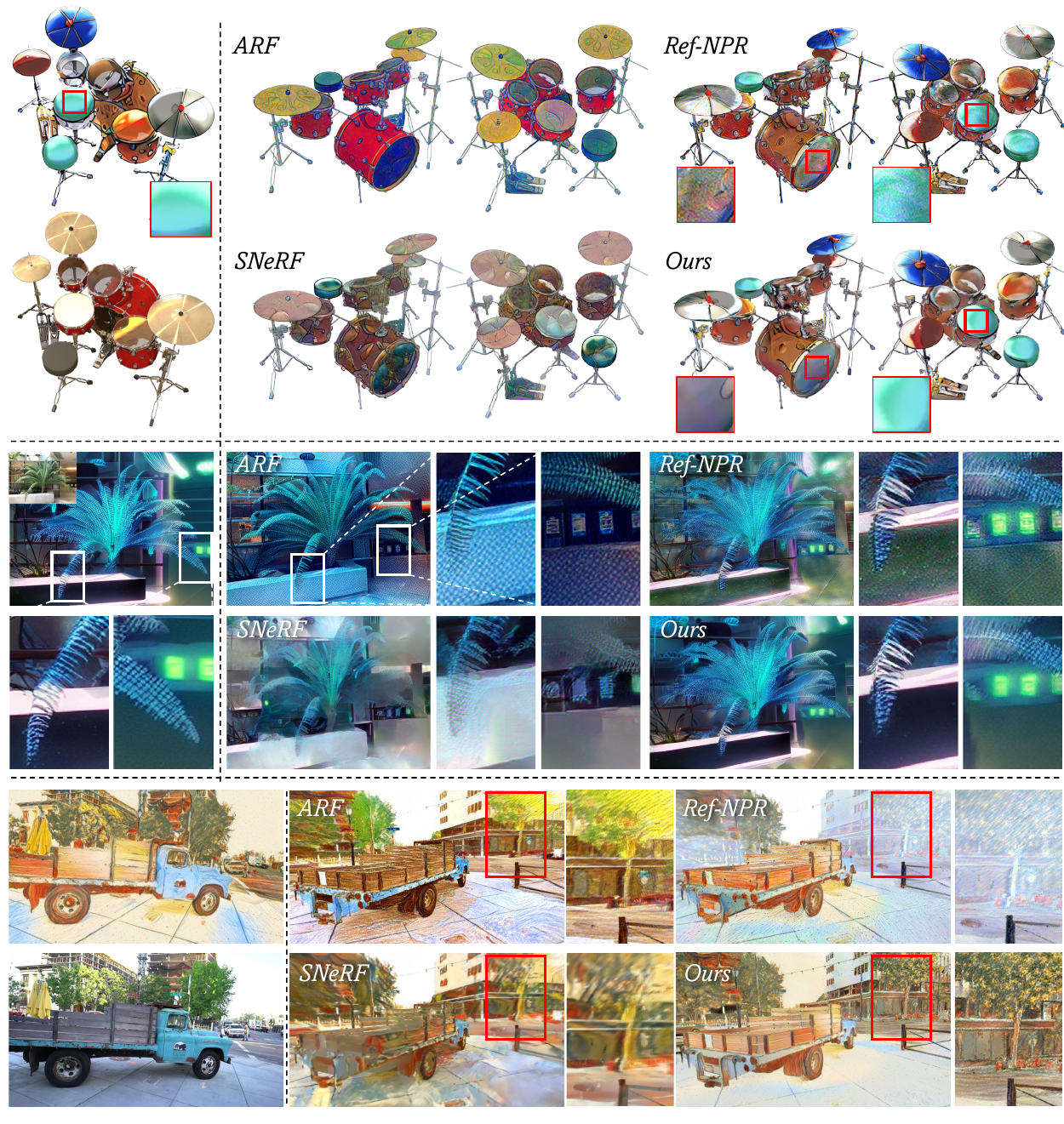}
\vspace{-8mm}	
 \caption{\textbf{Visual comparisons with state-of-the-art methods.} Paired reference and content view are shown on the left. Our method produces visual-compelling results that precisely follow the texture of the given reference, including the challenging high-frequency details such as the leaf in the second example. Baseline methods~\cite{ref-npr,arf,snerf} either lack semantic consistency or produce artifacts.}
 \vspace{-7mm}\label{llff}
\end{figure*}
\begin{table}[h]
\scriptsize
\centering
\caption{Quantitative comparison of different stylization methods.}
\vspace{-2mm}
\begin{tblr}{
  column{even} = {c},
  column{3} = {c},
  column{5} = {c},
  hline{1-2,5} = {-}{},
}
Metric      & ARF~\cite{arf}   & SNeRF~\cite{snerf} & Ref-NPR~\cite{ref-npr} & ReGS (Ours) \\
Ref-LPIPS$\downarrow$   & 0.394 & 0.405 & 0.339   & \textbf{0.202}  \\
Robustness$\uparrow$  & 26.34 & 26.03 & 28.11   & \textbf{31.27}     \\
Speed (fps) & 16.5  & 16.3  & 16.4    & \textbf{91.4}   
\end{tblr}
\vspace{-2mm}
\label{tab:quantitative}
\end{table}
\subsection{Compare with State-of-the-art Methods}
To evaluate stylization performance, we compare our method with three state-of-the-art baselines: ARF~\cite{arf}, SNeRF~\cite{snerf}, and Ref-NPR~\cite{ref-npr}. ARF~\cite{arf} and SNeRF~\cite{snerf} are general stylization methods that conduct style transfer without considering content correspondence. Ref-NPR~\cite{ref-npr} is a reference-based stylization method similar to ours that aims to precisely edit the 3D appearance based on the reference. All baselines are NeRF-based approaches built upon Plenoxels~\cite{plenoxels}.

\medskip
\noindent\textbf{Qualitative Evaluation.}
We report qualitative results in Figure~\ref{llff}. As shown, ARF~\cite{arf} and SNeRF~\cite{snerf} cannot generate semantic-consistent results with respect to the reference image as they ignore content correspondence. In contrast, Ref-NPR~\cite{ref-npr} produces more controllable results but yields artifacts. In some challenging cases (\eg last row in Figure~\ref{llff}), it also fails to achieve semantic consistent stylization (\ie green tree in the reference image is colored as white). In contrast, our method achieves better results that reproduce the desired texture, including challenging high-frequency ones.

\smallskip
\noindent\textbf{Quantitative Evaluation.} We present quantitative results in Table \ref{tab:quantitative}. Results are averaged over all scenes. We follow the protocol from~\cite{ref-npr} and report Ref-LPIPS and Robustness. Ref-LPIPS computes LPIPS~\cite{zhang2018unreasonable} score between the reference image and the 10 nearest test views. To calculate robustness, we first (1) train a stylized base model $m_{b}$ and use it to render a set of stylized views as new references; (2) then we use these references to train another set of stylized models and (3) compute PSNR results between images produced by them and $m_{b}$ (using the same camera path). To measure run-time efficiency, we also report run-time FPS on a single A5000 GPU. As shown in Table~\ref{tab:quantitative}, our method achieves the best results in terms of both quality and efficiency. Notably, our method enables real-time stylized view synthesis at 91 FPS.

\section{Conclusion}
In this work, we introduce \name, which adapts Gaussian Splatting for reference-based controllable scene stylization. \name adopts a novel texture-guided control mechanism to make high-fidelity appearance editing with ease. This is achieved by adaptively replacing responsible Gaussians with a denser set to express the desired appearance details. The control process is guided by texture clues for appearance editing while preserving original scene geometry through a depth-based regularization. We demonstrate the state-of-the-art scene stylization quality and effective designs of \name through extensive experiments. Benefiting from the high efficiency of 3DGS, our method naturally enables real-time stylized view synthesis. Discussions of limitations can be found in the supplemental file.

{\small
\bibliographystyle{unsrt}
\bibliography{nips}
}

\end{document}